\documentclass[letterpaper, 10 pt, conference]{ieeeconf}  
\IEEEoverridecommandlockouts              
\usepackage{dsfont}

\usepackage{mathtools}
\usepackage{duckuments}
\usepackage[T1]{fontenc}
\usepackage{tikz}
\usetikzlibrary{arrows}
\usepackage{graphicx}
\usepackage{xcolor}
\usepackage{bbm}
\usepackage{bm}
\usepackage{booktabs}
\usepackage{amssymb}
\renewcommand\fbox{\fcolorbox{white}{white}}
\usepackage{caption} 
\usepackage{amsmath}
\usepackage{nomencl}
\usepackage{algorithmic}
\usepackage{mathtools}
\usepackage{cuted}
\usepackage{blindtext}
\usepackage{scrextend}
\usepackage{booktabs,tabularx}
\usepackage[input-decimal-markers=.]{siunitx}
\usepackage{lipsum}
\usepackage{multicol}
\usepackage{soul}
\addtokomafont{labelinglabel}{\sffamily}
\usepackage{commath}
\IEEEaftertitletext{\vspace{-3\baselineskip}}
\usepackage{upgreek}
\usepackage[hyphens]{url}
\usepackage[hidelinks]{hyperref}
\hypersetup{breaklinks=true}
\usepackage{microtype}

\usepackage{titlesec}
\titlespacing*{\section}{0pt}{0.4ex plus 0.2ex minus 0.2ex}{0.2ex}
\titlespacing*{\subsection}{0pt}{0.35ex plus 0.2ex minus 0.2ex}{0.15ex}
\titlespacing*{\subsubsection}{0pt}{0.3ex plus 0.2ex minus 0.2ex}{0.1ex}

\AtBeginDocument{%
  \setlength{\jot}{2pt} 
}

\usepackage[font=footnotesize]{caption}

\title{\Large \bf 
3D-Printed Anisotropic Soft Magnetic Coating for Directional Rolling of a Magnetically Actuated Capsule Robot}

\author{Jin Zhou$^{1\dag}$, Chongxun Wang $^{1\dag}$, Zikang Shen $^{1}$ and Fangzhou Xia$^{1*}$
\vspace{1em}
\thanks{$^{1}$ The University of Texas at Austin, Walker Department of Mechanical Engineering, MINIMAX Lab, 78712, Austin, TX, USA}
\thanks{$^{*}$ Corresponding author: Fangzhou Xia, email: {\tt\small fangzhou.xia@austin.utexas.edu}}
\thanks{${\dag}$ These authors contributed equally to this work.} 
}

\begin{document}
\maketitle

\begin{abstract}
Capsule robots are promising tools for minimally invasive diagnostics and therapy, with applications from gastrointestinal endoscopy to targeted drug delivery and biopsy sampling. Conventional magnetic capsule robots embed bulky permanent magnets at both ends, reducing the usable cavity by about 10--20 mm and limiting integration of functional modules. We propose a compact, 3D-printed soft capsule robot with a magnetic coating that replaces internal magnets, enabling locomotion via a thin, functional shell while preserving the entire interior cavity as a continuous volume for medical payloads. The compliant silicone–magnetic composite also improves swallowability, even with a slightly larger capsule size. Magnetostatic simulations and experiments confirm that programmed NSSN/SNNS pole distributions provide strong anisotropy and reliable torque generation, enabling stable bidirectional rolling, omnidirectional steering, climbing on $7.5^{\circ}$ inclines, and traversal of 5 mm protrusions. Rolling motion is sustained when the magnetic field at the capsule reaches at least 0.3 mT, corresponding to an effective actuation depth of 30 mm in our setup. Future work will optimize material composition, coating thickness, and magnetic layouts to enhance force output and durability, while next-generation robotic-arm-based field generators with closed-loop feedback will address nonlinearities and expand maneuverability. Together, these advances aim to transition coating-based capsule robots toward reliable clinical deployment and broaden their applications in minimally invasive diagnostics and therapy.
\end{abstract}


\section{Introduction and Background}\label{sec_i}
Capsule robots are promising tools for minimally invasive diagnostics and therapy, with applications ranging from gastrointestinal endoscopy to targeted drug delivery and biopsy~\cite{Cao2024}. Among various actuation strategies, magnetic actuation is particularly attractive because it enables wireless control, deep tissue penetration, and safe operation under clinically acceptable field strengths~\cite{Kim2022ChemRev}. Over the past decade, advances in magnetic soft materials and manipulation systems have led to untethered microrobots capable of complex locomotion and targeted interventions in confined anatomical environments~\cite{Kim2018Nature, Zhang2021SciRobot, Lum2016PNAS}.

Most existing magnetic capsule robots embed large internal permanent magnets as their actuation sources~\cite{Yim2012TRO, Son2020SoftRobot}. While effective, this approach occupies a substantial fraction of the capsule’s limited interior, restricting the integration of medical modules such as cameras, sensors, or therapeutic payloads. In parallel, another line of research has focused on fully magnetic microrobots without a capsule structure, including helical swimmers and soft-bodied magnetic particles, which demonstrate agile locomotion in fluids but lack the capacity to encapsulate or transport functional payloads~\cite{Peyer2013Nanoscale, Zhang2021SciRobot}. These two strategies highlight a fundamental trade-off: embedding magnets enables capsule-based functions but reduces usable volume, while magnet-only microrobots maximize agility but sacrifice payload integration.

An alternative strategy is coating-based magnetic actuation, in which a thin functional shell around the capsule replaces the need for bulky internal magnets. This approach preserves the full internal cavity for functional modules~\cite{Kim2022ChemRev}, allows the use of biocompatible coatings, and can maintain reliable actuation performance over extended residence times in the gastrointestinal tract~\cite{Liu2021AFM, Zhang2021SciRobot}. Moreover, the coating design is compatible with programmable magnetic anisotropy achieved through controlled magnetization patterns~\cite{Kim2018Nature, Lum2016PNAS}, which can enable multimodal locomotion. Despite these advantages, systematic studies of coating-based capsule robots remain scarce, particularly with respect to rolling locomotion under realistic physiological conditions such as liquid environments, inclined surfaces, and obstacle-laden terrains.

This work introduces a compact, coating-based capsule robot with programmable magnetic anisotropy, and systematically evaluates its rolling locomotion under clinically relevant environments. Our main contributions are threefold:

\begin{itemize}
    \item \textbf{Design and simulation analysis:} We introduce a coating-based actuation mechanism that eliminates the need for internal permanent magnets and propose specific pole distributions (NSSN/SNNS) for rolling locomotion. Magnetostatic simulations and a simplified vector-alignment torque model~\cite{Peyer2013Nanoscale} are used to analyze anisotropy, predict torque generation, and guide design optimization.
    
    \item \textbf{Fabrication and continuous volume advantage:} We implement programmable anisotropy during 3D printing of a soft, biocompatible shell by controlling magnetization direction via G-code. In contrast to capsules with internal permanent magnets, which shorten the usable cavity by about 10 mm, the coating preserves the entire interior as a continuous volume, which also simplifies integration of medical components and provides greater flexibility for future modules. The compliant silicone–magnetic shell also improves swallowability, even with a slightly larger outer diameter.
    
    \item \textbf{Experimental validation and control:} We develop a vision-assisted testbed with controlled magnetic field modulation to evaluate locomotion. The capsule demonstrates stable bidirectional rolling, omnidirectional steering, incline climbing, and obstacle traversal, with reliable actuation at field strengths as low as 0.3~mT.
\end{itemize}



\newpage
\begin{strip}
\centering
\captionsetup{type=figure}     
\includegraphics[width=\textwidth]{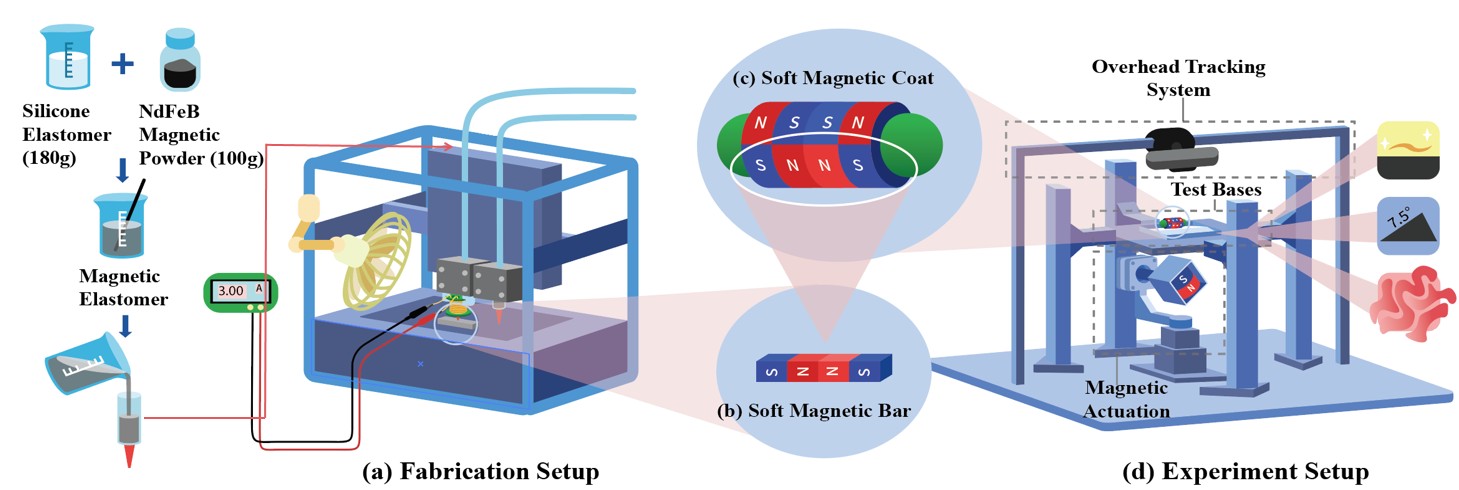}
\caption{Overview: (a) Fabrication Setup with 180\,g silicone + 100\,g NdFeB composite, (b) Soft magnetic bar with programmable polarity, (c) Capsule’s soft magnetic coat for rolling and directional rotating, (d) Experiment setup with diverse test bases.}
\label{fig:overview}
\label{fig:overview}
\end{strip}

This paper is organized as follows. Section~\ref{sec_ii} presents magnetostatic simulations of the coating design. Section~\ref{sec_iii} develops a dynamic model for rolling and yaw control. Section~\ref{sec_iv} describes the experimental setup and tracking method. Section~\ref{sec_v} reports locomotion results on different surfaces. Section~\ref{sec_vi} concludes with key findings and future directions.

\section{Overview}\label{sec_ii}
As illustrated in Fig.~\ref{fig:overview}, the experimental system is composed of several integrated subsystems. The overall design is centered on a bar magnet with an optimized pole distribution, which can be precisely controlled by external rotation. This bar serves as the basis for the capsule coating design, enabling accurate magnetic control while conserving the internal space of the capsule. 

For material fabrication, magnetic silicone with programmable pole distributions was prepared by mixing NdFeB magnetic powder with silicone elastomer. A coil was mounted around the 3D printer nozzle, and by synchronizing the nozzle position with G-code instructions, the current direction was dynamically adjusted to achieve programmable magnetic anisotropy during extrusion. 

The experimental evaluation consists of three main components: the magnetic actuation module, the test bases with different surface materials, and the overhead tracking system. The magnetic actuation module provides the primary driving force by generating and controlling the external magnetic field. The test bases are designed with various surface materials to assess the capsule’s locomotion stability under different conditions. Finally, the overhead tracking system records the trajectory of the capsule for subsequent analysis.

\section{Design and Analysis} \label{sec_iii}
\subsection{Magnetic Coating Design}
We designed a soft magnetic elastomer with an SNNS pole distribution, where two north poles are positioned at the center and two south poles are located at the ends. Since the two north poles are directly connected, the magnetic torque and force acting on each side are balanced. This configuration enables the generation of an overall central north polarity, with a stronger north polarity in the x-direction than in the z-direction because the bar is longer in the x-direction than in the z-direction. As a result, the x-direction of the bar is always attracted by the magnetic flux of the external magnet. Therefore, when the external magnetic field rotates, the bar also rotates synchronously in a horizontal plane, allowing for precise controlled movement of the bar.

Building upon the bar design, we extended the concept to the capsule coating. By patterning the upper half of the capsule as NSSN and the lower half as SNNS, the combined arrangement generates a net magnetic polarity oriented downward along the yaw axis of the capsule. As a result, when the external magnetic field rotates, the capsule experiences a corresponding torque that causes it to rotate synchronously, thereby enabling controllable orientation and directional movement.

\begin{figure}[htbp]
    \centering
    \includegraphics[width=1\linewidth]{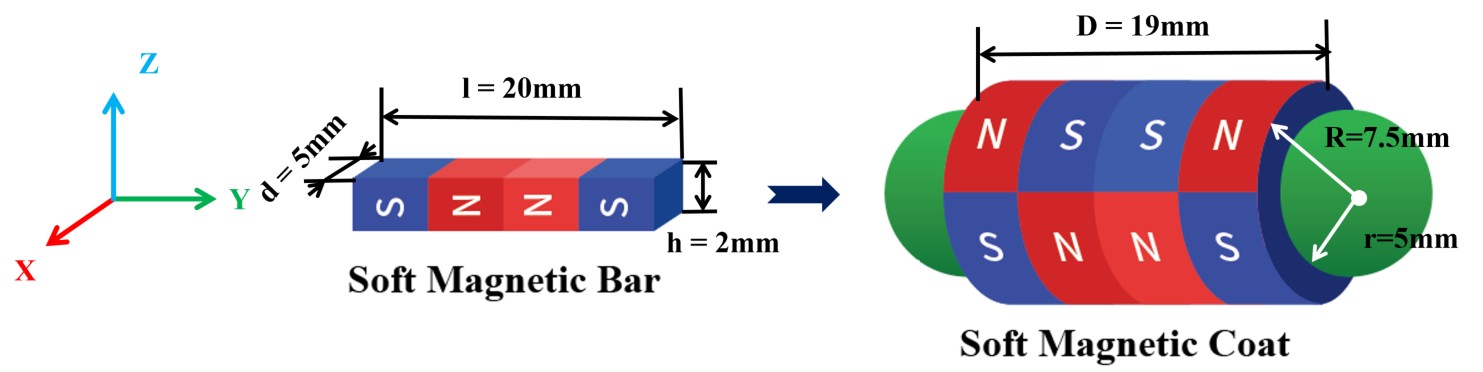}
    \caption{Magnetic coat design with outer radius R = 7.5\,mm, inner radius r = 5\,mm, and lenghth D = 19\,mm.}
    \label{fig:Design}
\end{figure}

The designed dimensions are illustrated in Fig.~\ref{fig:Design}. For the magnetic bar, the length is l = 20~mm, the width is d = 5~mm, and the height is h = 2~mm. For the magnetic coat, the outer radius is R = 7.5~mm, the inner radius is r = 5~mm, and the length is D = 19~mm, giving a volume of 1865.43~$mm^3$. The capsule itself has a volume of 1007.93~$mm^3$, meaning that the magnetic coat accounts for approximately 64.9\% of the total volume.
\vspace{-4mm}

\subsection{Simulation and Analysis} 
This section reports two magnetostatic simulations performed in Ansys Maxwell 3D to validate the magnetic field shaping logic of our magnetic coating design patterns using the 3D Magnetostatic solver.

\subsubsection{Modeling workflow (common to both studies)}\label{sec:sim:workflow}

\begin{figure}[htbp]
    \centering
    \includegraphics[width=\linewidth]{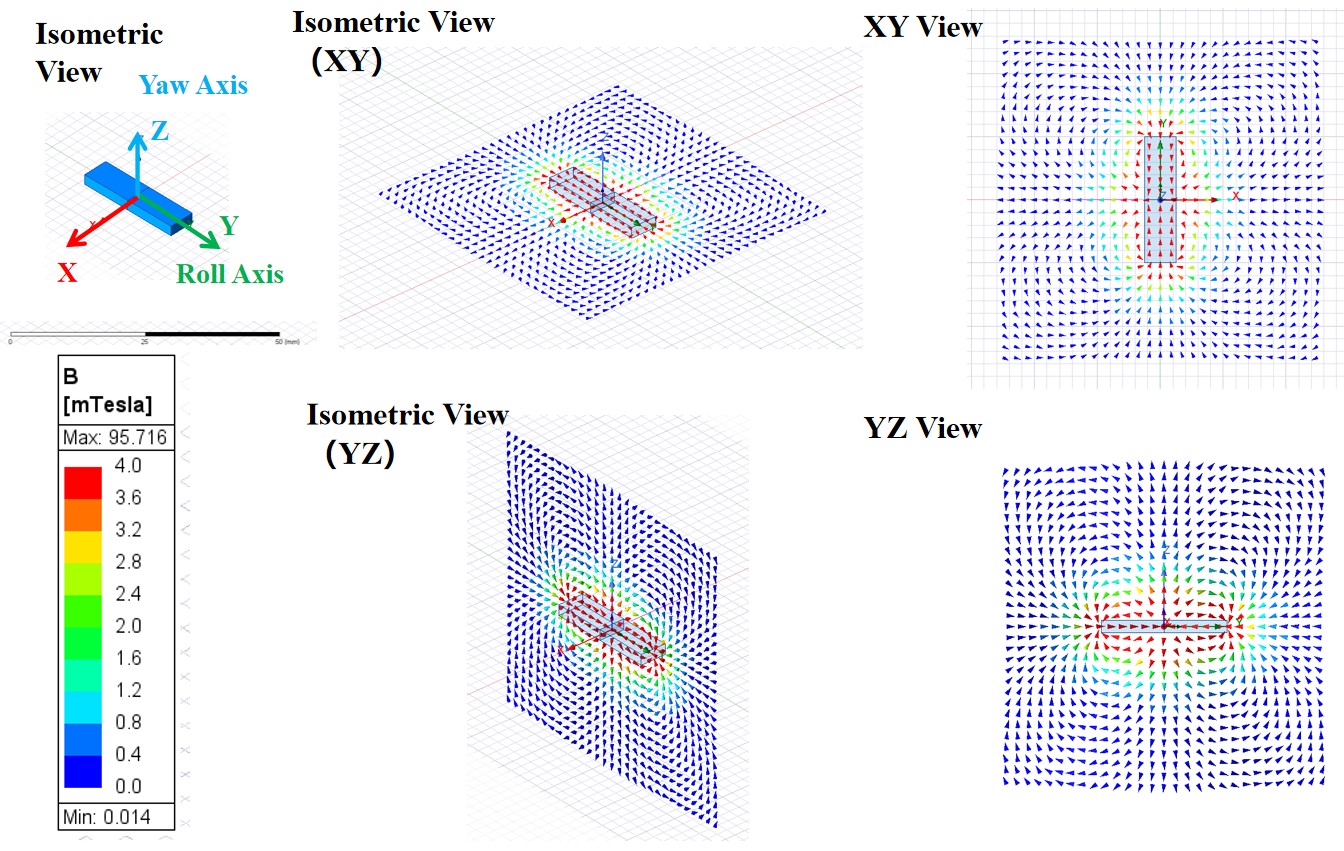}
    \caption{Simulation~I: "Basic bar" with \textsc{SNNS} pattern. Top: $B_z$ on $\Pi_z$; Bottom: $B_x$ on $\Pi_x$. $B_z$ is almost two times stronger than $B_x$.}
    \label{fig:bar_field}
\end{figure}

For both simulations, as shown in Fig.~\ref{fig:bar_field} and Fig.~\ref{fig:capsule_field}, the global coordinate system takes $z$ as vertical (yaw axis), $x$ as lateral (pitch axis), and $y$ as transverse (rolling axis). Elastomer structures were modeled at full scale in Ansys Maxwell 3D with assigned SNNS or NSSN magnetization patterns, each segment constrained to $\pm M_0 \hat{\mathbf{y}}$ while the substrate was treated as nonmagnetic. The models were embedded in an air region at least five times larger than the capsule dimensions, with a Zero-Tangential-H-Field boundary to approximate free space. A length–based mesh with maximum element size of 1~mm and a margin of 10~mm around the structure ensured adequate refinement. Field data were sampled on central planes: $B_x$, $B_y$, and $B_z$ over a $50 \times 50$~mm region for the bar, and over a $100 \times 100$~mm region for the capsule, with results exported for anisotropy analysis.

\vspace{-2mm}
\vspace{-5mm}

\begin{figure}[htbp]
    \centering
    \includegraphics[width=\linewidth]{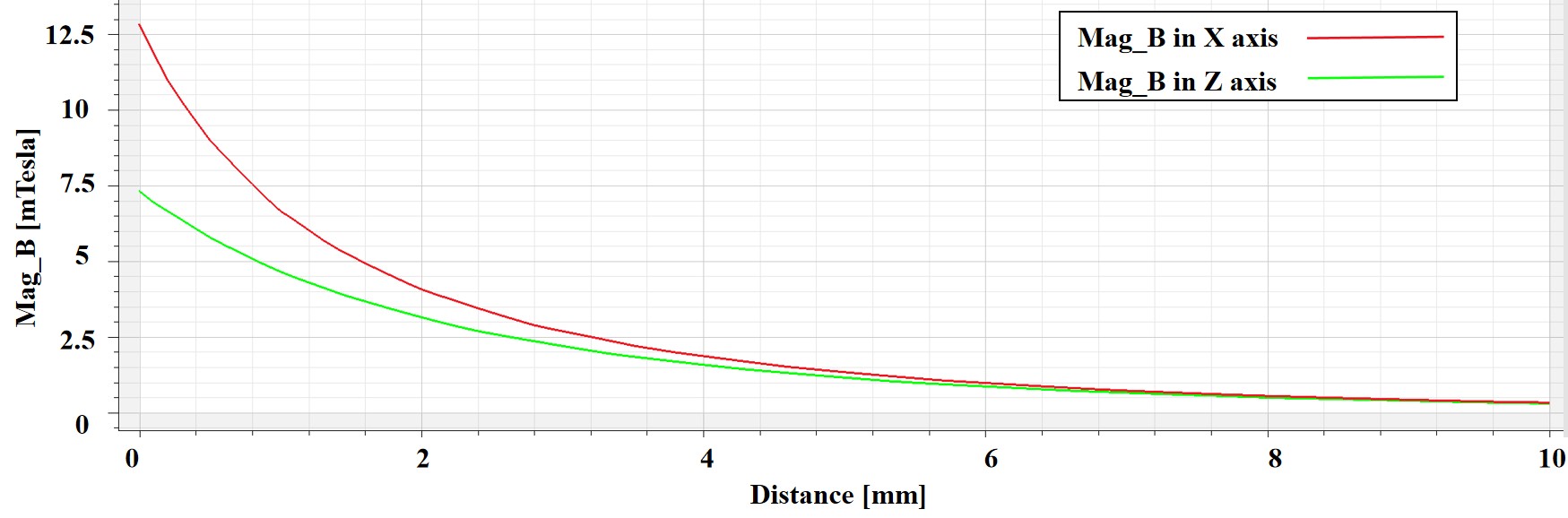}
    \caption{Magnetic field strength comparison between a line in the x-axis and a line in the z-axis (4mm to 14mm from the origin).}
    \label{fig:comparison}
\end{figure}


\subsubsection{Simulation~I: ``Basic bar'' (\textsc{SNNS}) and field anisotropy}\label{sec:sim:bar}
The first simulation examined a bar composed of elastomer segments arranged in the SNNS configuration, magnetized in $\pm \hat{\mathbf{y}}$ directions. Field maps on a $50 \times 50$~mm sampling plane (Fig.~\ref{fig:comparison}) showed that the $x$-axis component dominates the $z$-axis component, with $B_x$ nearly twice as strong as $B_z$ between 4--14~mm from the origin. This anisotropy indicates that the effective dipole moment of the bar is strongly aligned with the $x$-axis, ensuring predictable alignment with external fields. The bar design establishes a conceptual foundation for capsule orientation control, demonstrating how predictable and efficient rotation can be achieved.



\subsubsection{Simulation~II: Capsule coating (\textsc{NSSN}/\textsc{SNNS}) and resultant moment}\label{sec:sim:capsule}
\begin{figure}[htbp]
    \centering
    \includegraphics[width=\linewidth]{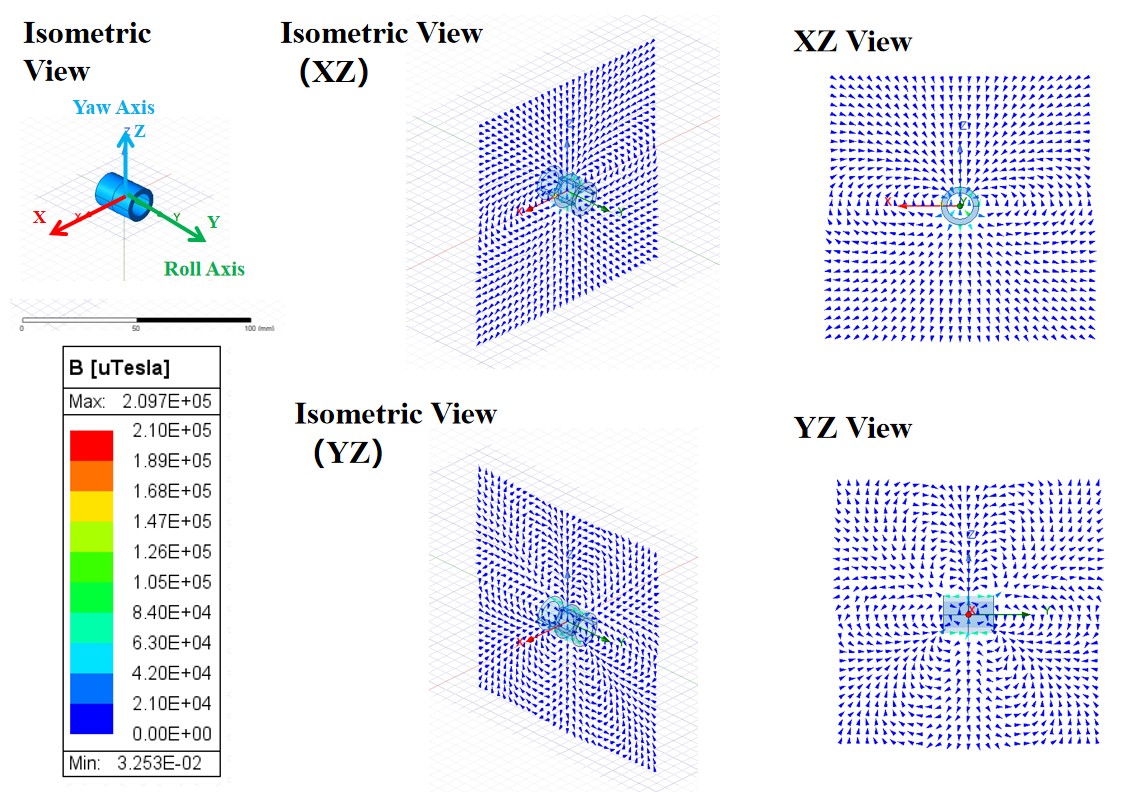}
    \caption{Simulation~II: with upper \textsc{NSSN} and lower \textsc{SNNS} shells. Streamlines and $B_z$ map show a net field oriented toward negative $z$.}
    \label{fig:capsule_field}
\end{figure}

The second simulation modeled a capsule coated with complementary pole patterns: the upper half in NSSN and the lower half in SNNS. Field sampling on $100 \times 100$~mm $xz$ and $yz$ planes (Fig.~\ref{fig:capsule_field}) showed that $B_z$ is significantly stronger than lateral components, producing a net polarity along the negative $z$-axis. This clear vertical moment ensures that under an actuation of an external rotating magnet, the capsule aligns its net magnetization with the applied field and rotates synchronously. The coating design thus enables reliable and capsule orientation control for rolling and steering.

\subsubsection{Insights and design implications}
The simulations confirm that the SNNS/NSSN configurations generate strong anisotropy and a clear net polarity, providing predictable alignment with external fields. These results explain why the chosen coating design is effective for orientation control. At the same time, they suggest possible extensions: increasing the number of pole segments (e.g., NSSNNSSN) could further enhance alignment torque, though at the cost of greater fabrication complexity. Overall, the current design balances anisotropy, manufacturability, and robustness, while leaving room for future optimization.

\subsection{Dynamics Modeling}
The dynamic model is based on the following assumptions:  
\begin{itemize}
    \item \textbf{Flat surface:} The capsule rolls on a continuous flat surface without geometric disturbances.  
    \item \textbf{No slip:} The capsule rolls without slip, so forward speed is kinematically linked to roll angle.  
    \item \textbf{Uniform field:} The external permanent magnet field is approximated as a uniform vector at the capsule location.  
    \item \textbf{Linearization:} Small misalignment angles are assumed, allowing the magnetic torque to be linearized as proportional to the angular difference between the external field direction and the capsule’s orientation.  
    \item \textbf{Decoupled DOFs:} Yaw and roll are treated as independent, neglecting cross-coupling.  
\end{itemize}

Under these assumptions, a minimal model captures two principal degrees of freedom (DOF): (i) yaw about the vertical axis that sets heading direction, and (ii) rolling about a lateral axis that drives forward translation. Consistent with simulation results, the capsule’s coating produces a net moment $\mathbf{m}$ aligned with its body axis and oriented predominantly along $-\hat{\mathbf{z}}$. The external field $\mathbf{B}_{\mathrm{ext}}(t)$ is provided by rotating magnets with controllable orientation and approximately constant magnitude $B=\|\mathbf{B}_{\mathrm{ext}}\|$ in the workspace.


\subsubsection{Free-body diagram and coordinates}
Let $\mathcal{W}=\{x,y,z\}$ denote the world frame. The generalized coordinates are:  
\begin{itemize}
    \item $\psi$ (yaw): rotation of the capsule about $\hat{\mathbf{z}}$.  
    \item $\alpha$ (roll): rotation about a lateral axis; with radius $R$, the no-slip constraint gives $v = R\,\dot{\alpha}$.  
\end{itemize}

\begin{figure}[htbp]
    \centering
    \includegraphics[width=1\linewidth]{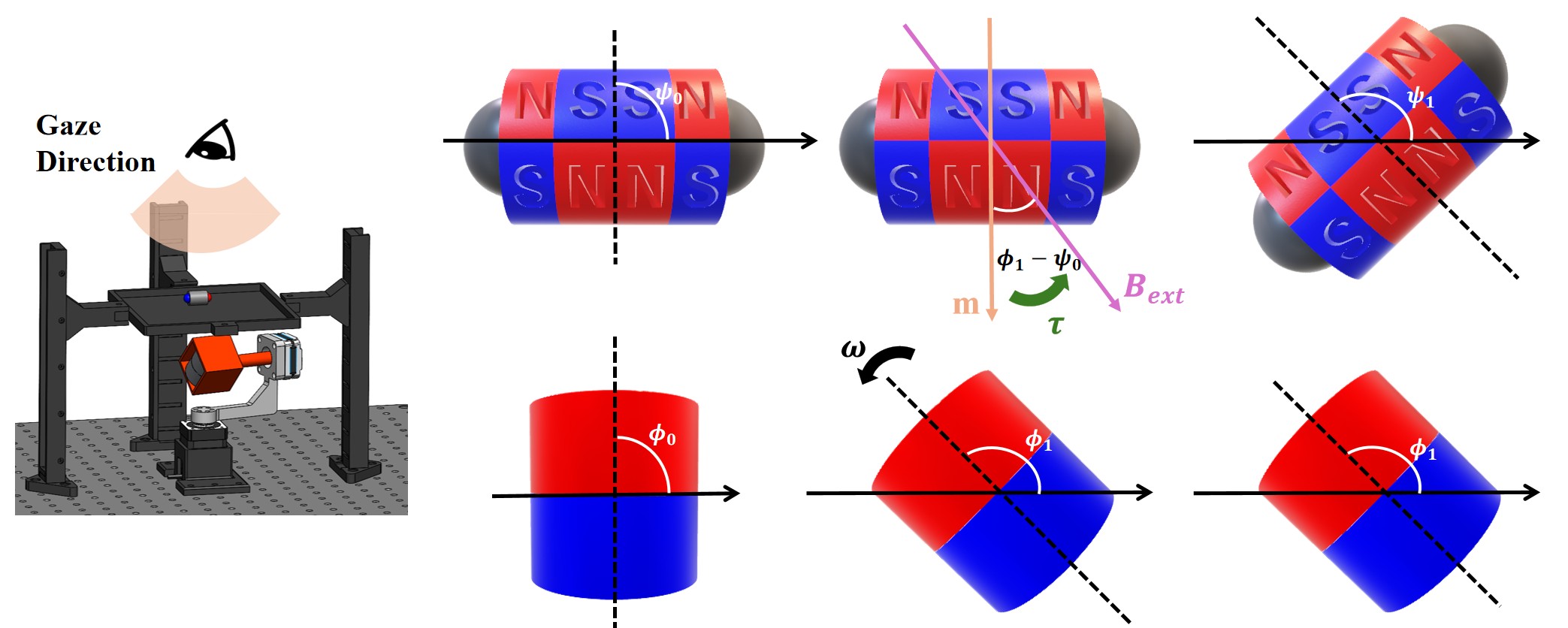}
    \caption{Top view of the experiment setup showing the yaw motion of the capsule caused by the rotation of the external magnet about z-axis. The capsule has net moment $\mathbf{m}$; the external field $\mathbf{B}_{\mathrm{ext}}$ is commanded by azimuth $\phi$. Dominant torque components excite yaw $\psi$.}
    \label{fig:top}
\end{figure}

\begin{figure}[htbp]
    \centering
    \includegraphics[width=1\linewidth]{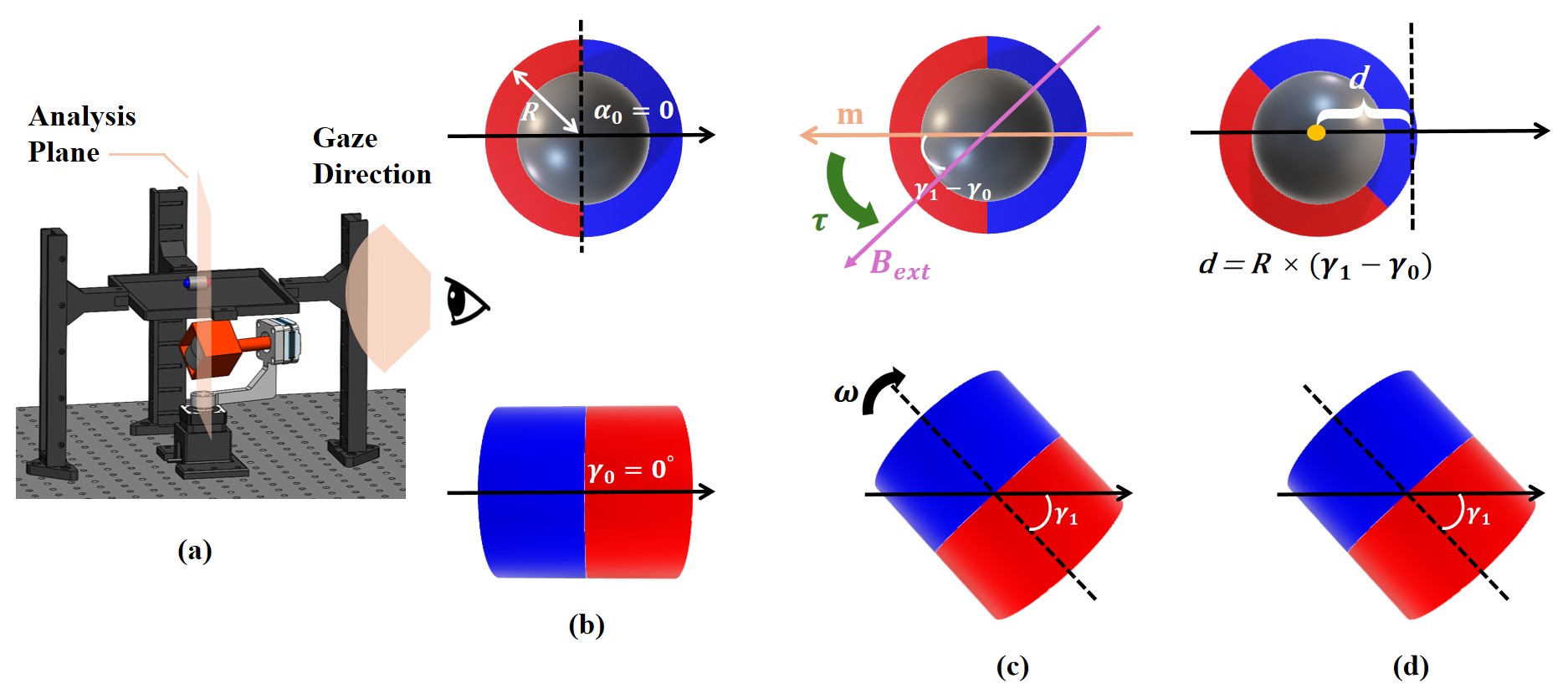}
    \caption{Side view of the experiment setup showing the roll motion of the capsule caused by the rotation of the external magnet about y-axis. The capsule has net moment $\mathbf{m}$; the external field $\mathbf{B}_{\mathrm{ext}}$ is commanded by tilt $\gamma$. Dominant torque components excite roll $\alpha$.}
    \label{fig:side}
\end{figure}

As shown in Fig. \ref{fig:top} and Fig. \ref{fig:side}, the external field orientation is parameterized by two control angles: azimuth $\phi(t)$ in the $x$--$y$ plane and tilt $\gamma(t)$ orthogonal to $\hat{\mathbf{z}}$. The magentic torque is
\begin{equation}
\boldsymbol{\tau}_m = \mathbf{m}\times\mathbf{B}_{\mathrm{ext}} \, ,
\end{equation}
with dominant components $\tau_{m,\phi}$ driving yaw $\psi$ and $\tau_{m,\gamma}$ driving roll $\alpha$.

\vspace{-5mm}

\vspace{-3mm}
\subsubsection{Equations of motion (2-DOF)}
Each DOF is modeled as a rigid-body rotation with inertia and viscous drag. For small misalignment and decoupled motion, the magnetic torque linearizes as $\tau_m \approx mB(\text{command angle} - \text{capsule angle})$. Including damping $c_\psi, c_\alpha$ and inertias $I_\psi, I_\alpha$, the governing equations are
\vspace{-1mm}
\begin{align}
I_\psi \ddot{\psi} + c_\psi \dot{\psi} &= mB\,(\phi - \psi), \label{eq:yaw_eom}\\
I_\alpha \ddot{\alpha} + c_\alpha \dot{\alpha} &= mB\,(\gamma - \alpha) - \tau_{\text{load}}(\dot{\alpha}), \label{eq:roll_eom}
\end{align}

where $\tau_{\text{load}}$ represents rolling contact effects (substrate slope, etc.). For motion above the static-friction threshold, $\tau_{\text{load}} \approx c_f\dot{\alpha}$ can be absorbed into $c_\alpha$.

Rolling begins when $mB \ge \tau_s$, with $\tau_s \approx \mu N R$ (normal load $N$, friction coefficient $\mu$, and capsule radius $R$), giving a minimum actuation field $B_{\min} \approx \tau_s/m$.

\subsubsection{Transfer functions and block diagram}
Taking Laplace transforms of \eqref{eq:yaw_eom}--\eqref{eq:roll_eom} with zero initial conditions yields
\vspace{-1mm}
\begin{align}
\frac{\Psi(s)}{\Phi(s)} &= \frac{mB}{I_\psi s^2 + c_\psi s + mB}
= \frac{\omega_{n,\psi}^2}{s^2 + 2\zeta_\psi \omega_{n,\psi} s + \omega_{n,\psi}^2}, \\
\frac{A(s)}{\Gamma(s)} &= \frac{mB}{I_\alpha s^2 + c_\alpha s + mB}
= \frac{\omega_{n,\alpha}^2}{s^2 + 2\zeta_\alpha \omega_{n,\alpha} s + \omega_{n,\alpha}^2},
\end{align}


with natural frequencies and damping ratios
\[
\omega_{n} = \sqrt{\tfrac{mB}{I}}, \qquad 
\zeta = \tfrac{c}{2\sqrt{ImB}} .
\]

\begin{figure}[htbp]
    \centering
   \includegraphics[width=1\linewidth]{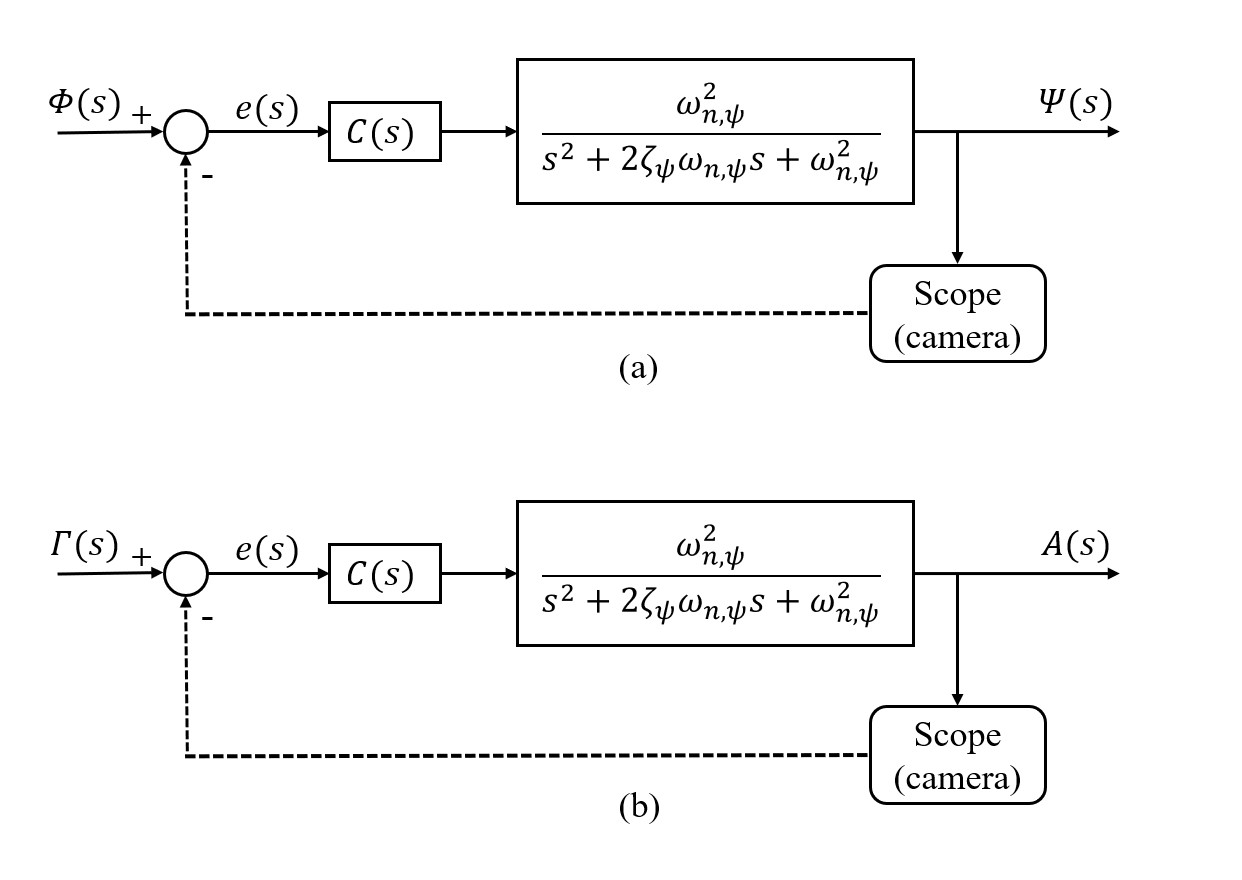}
    \caption{Open-loop block diagram of the two-DOF capsule system. Magnetic field orientation $(\phi,\gamma)$ drives the capsule dynamics, producing yaw and roll. A camera provides measurement but no feedback.}
    \label{fig:blockdiagram}
\end{figure}

Both yaw and roll thus behave as second-order low-pass trackers of the commanded field angles $(\phi,\gamma)$. Figure~\ref{fig:blockdiagram} illustrates the open-loop mapping: magnetic inputs $(\phi,\gamma)$ drive the capsule dynamics, producing yaw $\psi$ and roll $\alpha$. A camera measures orientation and position for validation, but no feedback path is used.
\vspace{-5mm}

\subsubsection{Quasi-static correspondence of magnet angle and motion}
In the quasi-static regime, the magnetic stiffness term $mB$ acts like a torsional spring that drives alignment between the capsule’s net moment and the external field. As a result, the capsule asymptotically tracks the commanded field angle:
\begin{equation}
    \lim_{t\to\infty}\frac{\psi(t)}{\phi(t)} = \lim_{s\to 0}\frac{\Psi(s)}{\Phi(s)} = 1, 
~~
\lim_{t\to\infty}\frac{\alpha(t)}{\gamma(t)} = \lim_{s\to 0}\frac{A(s)}{\Gamma(s)} = 1 .
\end{equation}

Using simulation estimates of m = 0.024-0.093~A$\cdot$ $m^2$ and B = 0.0308~T, with inertias $I_\psi$ = 3.74$\times10^{-7}$~kg$\cdot m^2$ and $I_\alpha$=2.06$\times10^{-7}$~kg$\cdot m^2$, the undamped natural frequencies are $f_{n,\psi} \approx 7.1\text{--}13.9~\mathrm{Hz}, ~
f_{n,\alpha} \approx 9.6\text{--}18.8~\mathrm{Hz}$.

In experiments, the external magnet was rotated at $60^{\circ}$/s for rolling (0.17 Hz) and $4^{\circ}$/s for yaw (0.011 Hz), both far below these values. Since damping was neglected in the estimates, the true resonant frequencies would be slightly lower. In practice, rolling rates of external magnet is significantly lower than the resonance, which ensures that capsule operates in the quasi-static regime with negligible phase lag and accurate tracking of commanded field angles.


\subsubsection{Command mapping to planar position}
Planar locomotion can be decomposed into sequential yaw and roll commands. To move toward a target $(X,Y)$, the capsule first yaws by $\psi$ to align with the displacement vector, then rolls by $\alpha$ to generate translation without slip.
\begin{equation}
\Delta s = R\,\Delta\alpha
\end{equation}
links commanded roll to forward displacement $\Delta s$, where $R$ is the capsule radius. This simple yaw-then-roll mapping enables open-loop point-to-point control without simultaneous multi-DOF actuation.

\subsubsection{Model limitations and parameter estimation}
The model relies on several simplifying assumptions. The external permanent magnet field is approximated as a uniform vector at the capsule location, which is reasonable when the capsule operates within a limited workspace far from the magnet where gradients are small. Yaw and roll dynamics are treated as decoupled and valid for small misalignment angles; cross-coupling becomes significant only at large tilts. Damping coefficients $c_\psi$ and $c_\alpha$ are modeled as linear viscous terms, which approximate rolling resistance but neglect stick–slip and anisotropic friction.

With estimated parameters m = 0.024-0.093~A$\cdot m^2$, B = 0.0308~T, $I_\psi$=3.74$\times10^{-7}$~kg$\cdot m^2$, and $I_\alpha$ = 2.06$\times10^{-7}$~kg$\cdot m^2$, the undamped natural frequencies are $f_{n,\psi}$ = 7.1-13.9 Hz and $f_{n,\alpha}$ = 9.6-18.8 Hz. These values can be refined using step or sine-sweep tests, while damping can be identified from transient decay. The magnetic gain mB is obtained directly from the simulated moment and applied field. 

Although damping and nonlinearities would reduce the true resonant frequencies slightly, the large separation from experimental actuation rates (0.17 Hz roll, 0.011 Hz yaw) ensures quasi-static operation. Future work can incorporate feedback control to address higher-frequency dynamics and expand operation beyond this regime.

\section{Experimental Setup}\label{sec_iv}
\begin{figure}[htbp]
    \centering
    \includegraphics[width=1\linewidth]{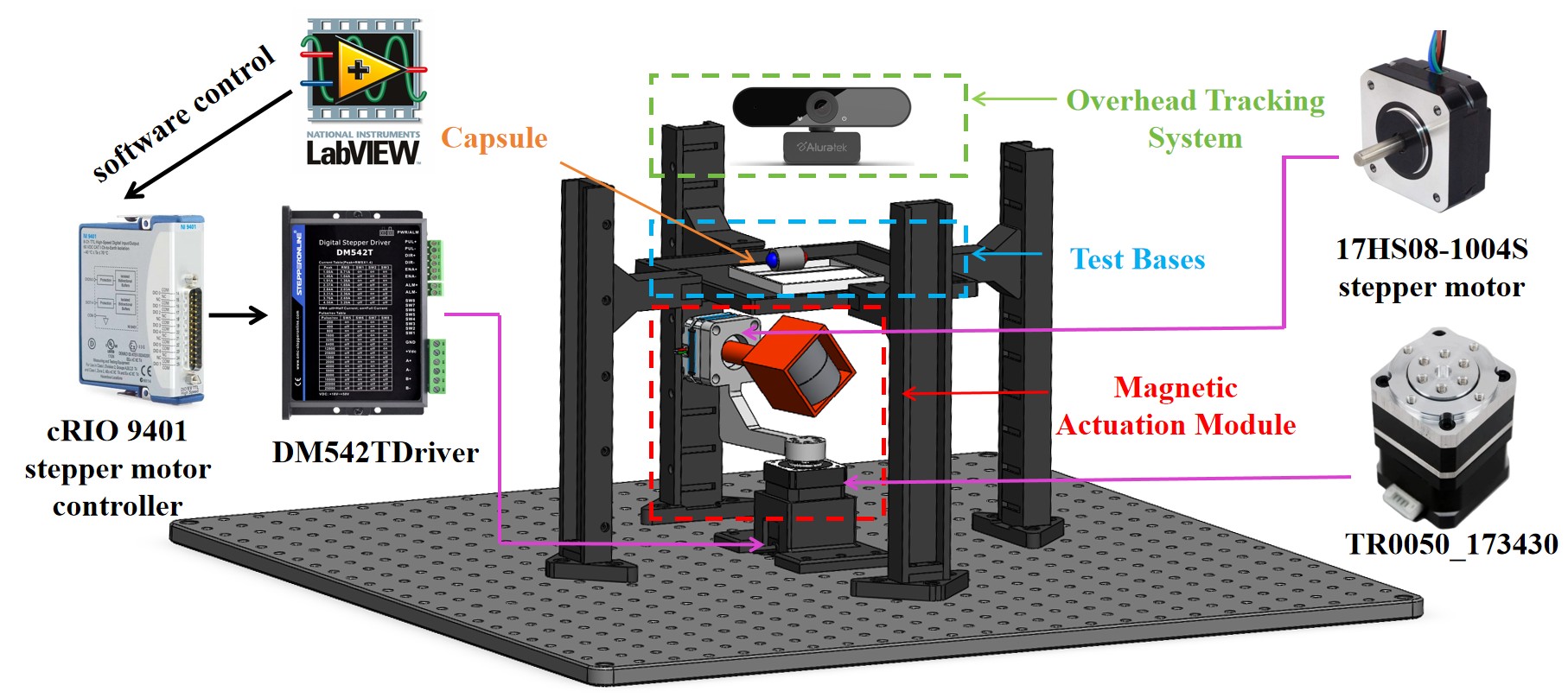}
    \caption{Overview of the experimental setup.}
    \label{fig:setup}
\end{figure}

As shown in Fig.~\ref{fig:setup}, the experimental setup consists of four main parts: (1) the capsule with magnetic coating, (2) the magnetic actuation module, (3) the test bases with different surface materials, and (4) the overhead tracking system. 

\subsection{3D Printer for Capsule Magnetic Coating Fabrication}
In order to realize the capsule robot design, a soft magnetic coating is fabricated that combines mechanical flexibility with programmable magnetization. To prepare the magnetic elastomer, we adopted a 1.8:1 weight ratio of neodymium--iron--boron (NdFeB) magnetic powder and silicone elastomer (hardness: 18~Shore A) ~\cite{liu2022reprogrammable}. The NdFeB magnetic powder was purchased from Nanochemazone Inc. (Canada; catalog NCZM-125-19, particle size $\sim$50~µm, purity $>$99.9\%), and the silicone elastomer was sourced from Sandraw. This composition provides both strong magnetization and mechanical flexibility suitable for printing.

The printing was carried out using a Sandraw S180 3D printer, which allows deposition of the silicone–magnetic composite. The principle of programmable magnetic anisotropy relies on aligning magnetic domains under a controlled external field, which enables spatially defined magnetization patterns~\cite{Kim2018Nature}. In our setup, a 100-turn copper coil was mounted around the printer nozzle and driven by a regulated 3~A current from a DC power supply, producing a magnetic field of approximately 19~mT. By reversing the current direction according to G-code instructions, the polarity of the deposited magnetic silicone could be dynamically programmed. An electric cooling fan was added to dissipate heat from the coil and prevent overheating of the printhead assembly. We adopted a 0.4~mm nozzle, selecting a relatively large orifice to prevent clogging by the magnetic silicone, which tends to solidify more quickly under heating.

This method enables the fabrication of silicone structures with arbitrary shapes and customized magnetic domain distributions, providing a reproducible and versatile platform for subsequent locomotion experiments.

\begin{figure}[htbp]
    \centering
    \includegraphics[width=1\linewidth]{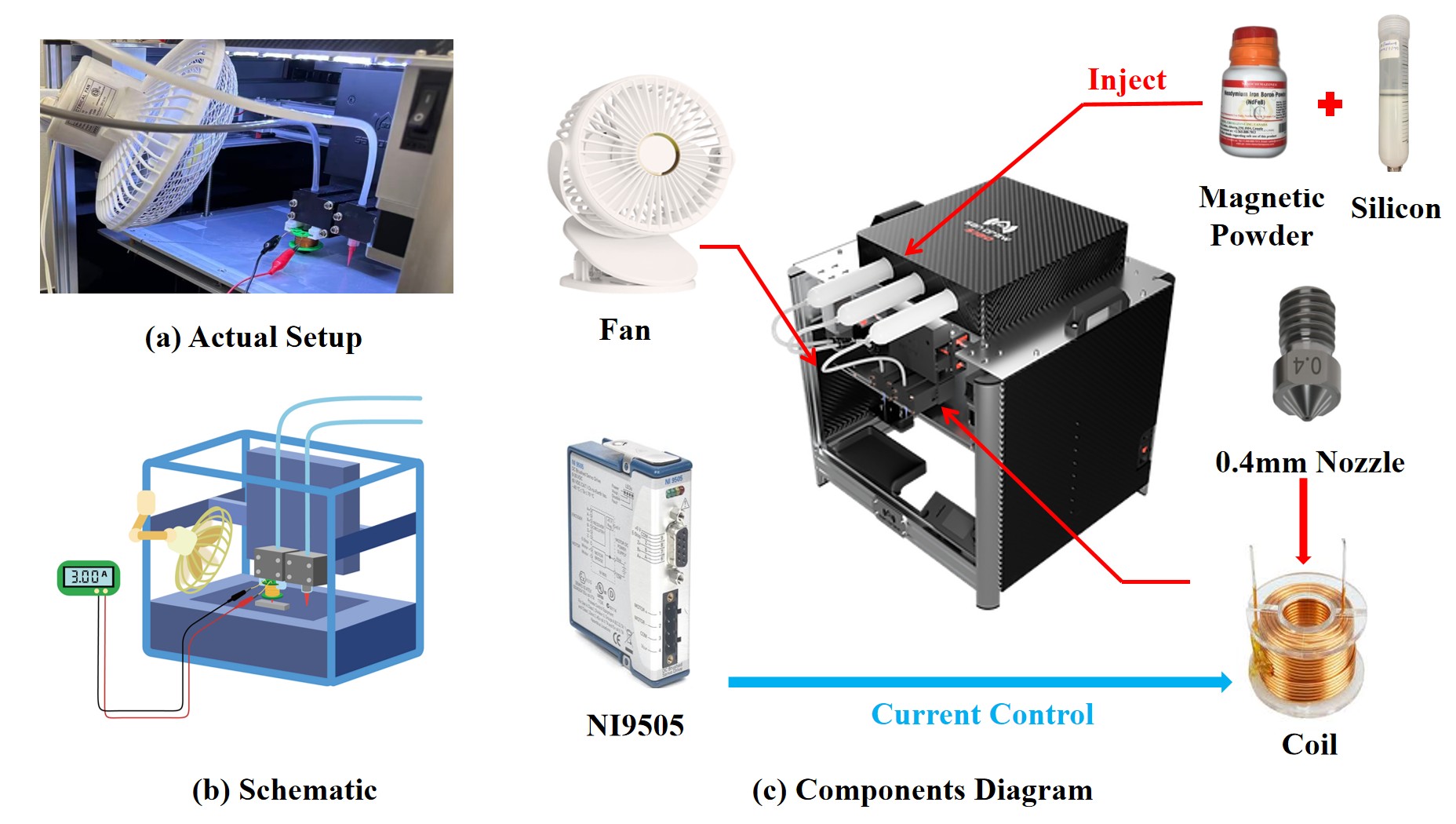}
    \caption{3D printing system: (a) actual setup; (b) schematic; (c) components siagram.}
    \label{fig:fabrication}
\end{figure}

\subsection{Magnetic Actuation Module}
\vspace{-1mm}
The actuation system consists of two stepper motors and a permanent magnet. One motor (17HS08-1004S), a two-phase stepper motor with a step angle of 1.8°, controls the rotation of the magnet about the horizontal axis to drive the capsule forward or backward via rolling. The other motor (TR0050-173430), also a two-phase stepper motor but with a fine step angle of 0.06°, controls rotation about the vertical axis to change the capsule's direction of motion. This configuration enables independent control of propulsion and steering. Both stepper motors are controlled through a LabVIEW program running on a CompactRIO controller (NI-9074) equipped with a NI-9401 digital I/O module. The FPGA module in the controller operates with a 40 MHz clock, ensuring high-precision timing for motor commands. By sending a move command every 300,000 counts to the TR0050-173430 motor, the bottom stepper motor rotates the magnet about the vertical axis at 4°/s. Similarly, by sending a move command every 600,000 counts to the 17HS08-1004S motor, the horizontal stepper motor achieves a rotation speed of 60°/s. Together, these elements allow for precise rotational speed control of the permanent magnet underneath the testbed, enabling accurate magnetic actuation along both horizontal and vertical axes.
\subsection{Test Bases with Different Surface Materials}
As summarized inTable~\ref{tab:test_bases}, we designed four test bases, inspired by gastric texture and geometry, to provide quasi-realistic locomotion tests. Normal stomach wall thickness is 2-4~mm in the body and 3-5~mm in the pylorus\cite{karnul2022study}; longitudinal gastric folds vary with distention and are typically  4~mm thick\cite{rapaccini1988gastric,susmallian2022correct}. Accordingly, our testbed includes silicone bases with 2~mm and 5~mm modular protrusions and a wet surface to mimic a fluid environment. The magnet–testbed spacing was set to 20~cm. The smooth base was 3D-printed in PLA, while compliant bases were 3D-printed in silicone (Shore A 18) to approximate gastric compliance and to realize inclines and folds. All silicone bases share identical dimensions (75~mm $\times$ 75~mm) and were mounted on the PLA platform.
\begin{table}[htbp]
\centering
\begin{tabular}{|p{2.7cm}|p{4.65cm}|}
\hline
\textbf{Base Type} & \textbf{Features} \\ \hline
\textbf{Smooth PLA Base} & Flat, low-friction surface \\ \hline
\textbf{Silicone Base with Slope} & Inclined surface with $10^\circ$ on each side \\ \hline
\textbf{Dry Silicone Base with Protrusions} & Inclined surface with 2~mm protrusions in the middle and 5~mm protrusions on the edge\\ \hline
\textbf{Wet Silicone Base with Protrusions} & Inclined surface with protrusions and water \\ \hline
\end{tabular}
\caption{Summary of the test bases and their features for experiments.}
\label{tab:test_bases}
\end{table}

\subsection{Overhead Capsule Tracking System}
\vspace{-1mm}
Real-time tracking of the capsule’s motion is performed using a Logitech MX Brio 960 webcam mounted above the test surface. The camera records at a resolution of $1920\times1080$ pixels and 30~Hz, which corresponds to a spatial detection accuracy of approximately 0.1~mm on the testbed. A Python program processes the video stream to identify red and blue markers affixed to opposite ends of the capsule via RGB thresholding. The capsule’s center position is calculated as the midpoint of the two detected centroids, while its orientation angle is determined from the line connecting them. To calibrate the camera frame to the testbed coordinate system, a 3D-printed fixture with five red dots at known $(x,y)$ positions is used. The detected pixel coordinates of these dots are then mapped to physical surface coordinates through a transformation matrix, enabling accurate measurement of capsule position and orientation in real-world units.

\section{Result and Discussion}\label{sec_v}
\vspace{-1mm}
In this section, we present the results of the capsule robot locomotion experimenton different surfaces.
\subsection{Rectangular Bar Locomotion on Smooth Surface}
\vspace{-1mm}
The rectangular bar was tested on a flat surface to assess its rolling-based locomotion under rotating magnetic fields. The bar’s heading is controlled by rotating the external magnet about its longitudinal axis, keeping the bar oriented perpendicular to the magnetic flux lines. Forward rolling is achieved when the magnet rotates about the transverse axis, with the bar continuously aligning itself perpendicular to the flux. During rolling, slight deviations in yaw can occur, likely due to surface unevenness or nonuniform magnetization from the printing process. However, because the bar naturally reorients to minimize misalignment with the field, these yaw deviations are corrected over time, allowing the bar to return to its original heading. This principle enables precise control of the bar’s movement, with only minor transient fluctuations.
\vspace{-2mm}

\subsection{Capsule Locomotion on Smooth 3D-printed Surface}
\vspace{-1mm}
 The robot demonstrated translational and rotational movement through the rotation of the permanent magnet underneath around its horizontal and vertical axes.

\begin{figure}[htbp]
    \centering
    \includegraphics[width=0.95\linewidth]{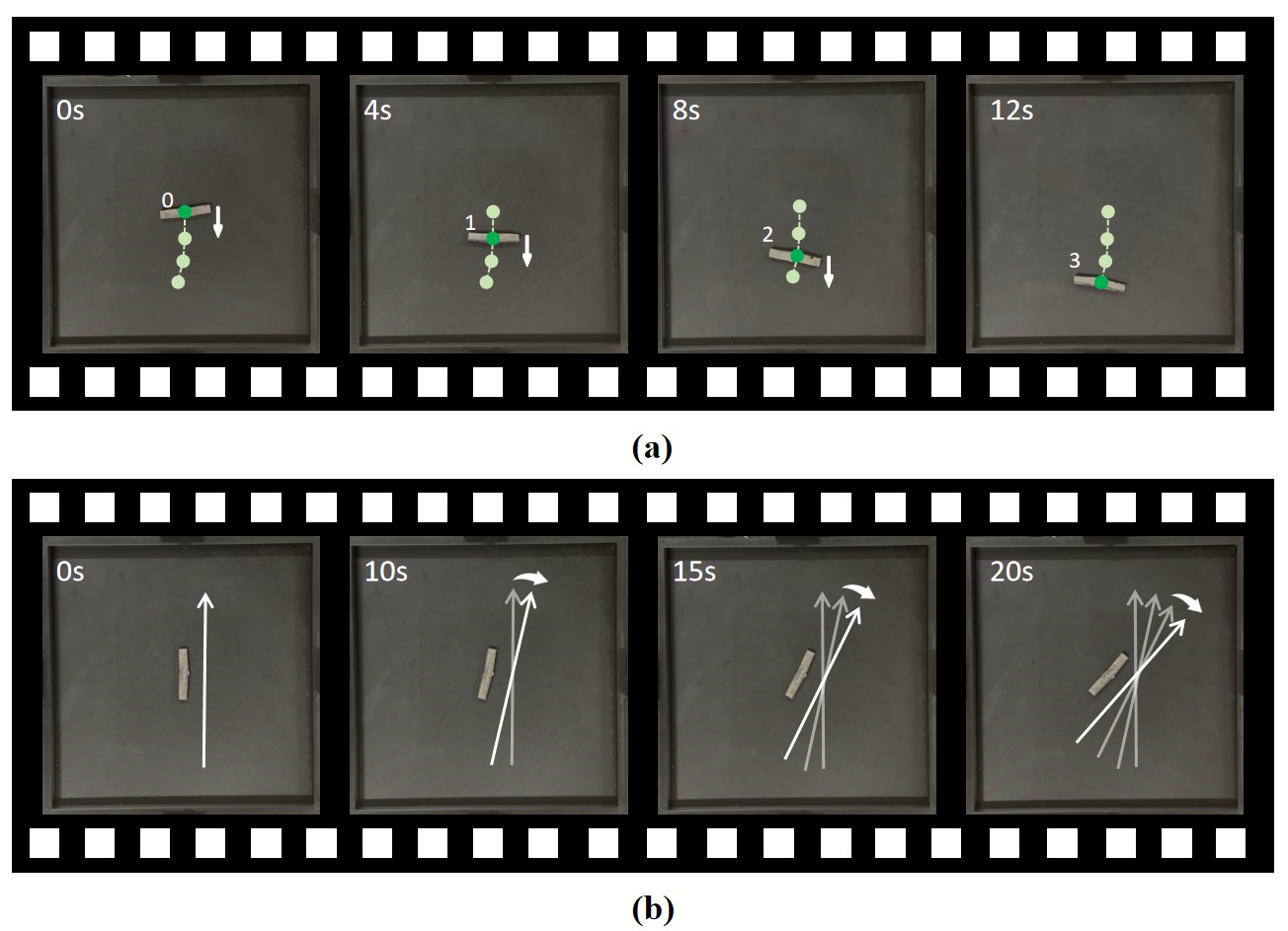}
    \caption{Movement of megnetic bar: (a) Bar rolling sequence at every 2-second intervals; (b) Bar rotation at every 5-second intervals.}
    \label{fig:roll_sequence}
\end{figure}
\vspace{-2mm}
 
\subsubsection{Continuous Roll}
The capsule exhibited smooth forward rolling as it translated across the flat 3D printed surface. It went steadily with minimal deviation from its expected path. In this rolling mode, the external magnet does not need to be directly beneath the capsule; we observed reliable propulsion with a lateral offset up to  30~mm, at which the local field strength at the capsule is 0.3~mT. When the stepper motor rotated around the horizontal axis at a speed of 10~rpm, the capsule achieved a forward translation speed of approximately 12.5~mm/s. A relatively greater discrepancy of motion can be observed as the permanent magnet rotates faster, potentially caused by inertial lag or slipping on the surface. Sequential positions were extracted at constant intervals and are shown in Fig.~\ref{fig:roll_sequence}.
\vspace{-2mm}
\begin{figure}[htbp]
    \centering
    \includegraphics[width=0.95\linewidth]{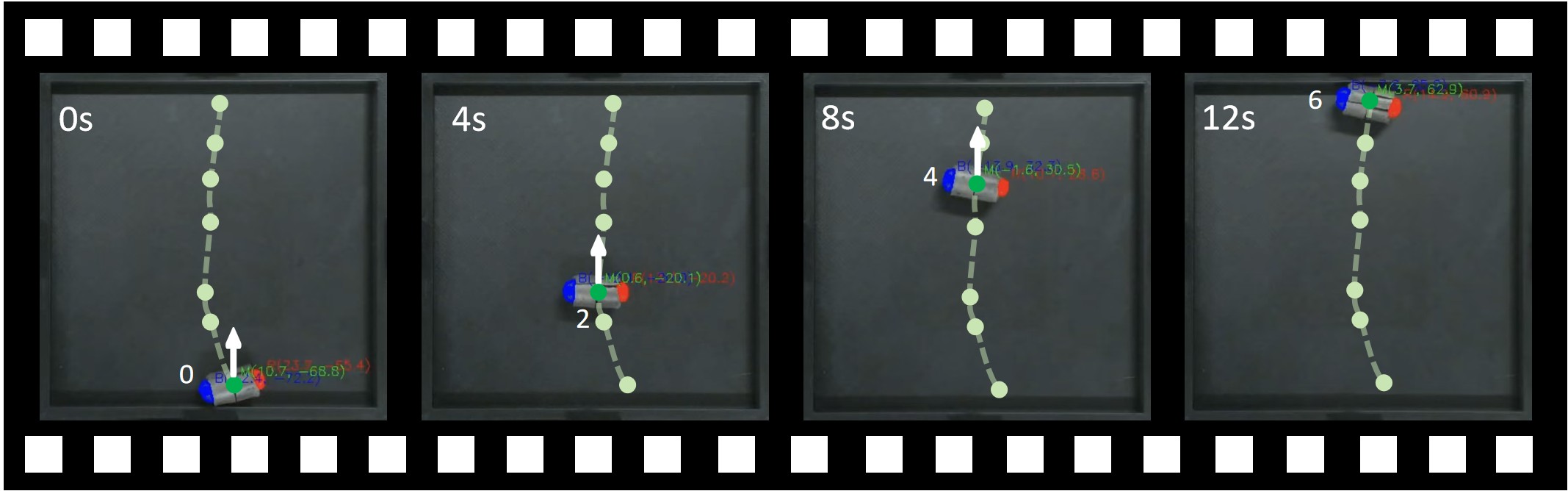}
    \caption{Capsule rolling sequence at 4-second intervals.}
    \label{fig:roll_sequence}
\end{figure}
\vspace{-2mm}

\subsubsection{Yaw angle control}
The capsule’s heading was controlled by rotating the external magnet about the vertical axis. At a relatively slow vertical-axis rotation of 0.67~rpm, the capsule’s yaw angle remained phase-locked to the magnet’s rotation, enabling smooth and consistent turning under low-friction conditions. Robust yaw control, however, required the magnet to be positioned directly beneath the capsule, with lateral offset tolerances of no more than $2~\mathrm{mm}$; exceeding this limit resulted in phase slips and significant heading errors. Because only along the vertical line beneath the capsule is the local magnetic field at the capsule approximately horizontal; lateral offsets introduce a vertical field component that generates a rolling torque, coupling yaw with roll and disrupting pure heading control. In the future, controlled lateral and axial motion of the permanent magnet, for instance using a robotic arm, will be incorporated to maintain precise alignment during in-body navigation.

\begin{figure}[htbp]
    \centering
    \includegraphics[width=0.95\linewidth]{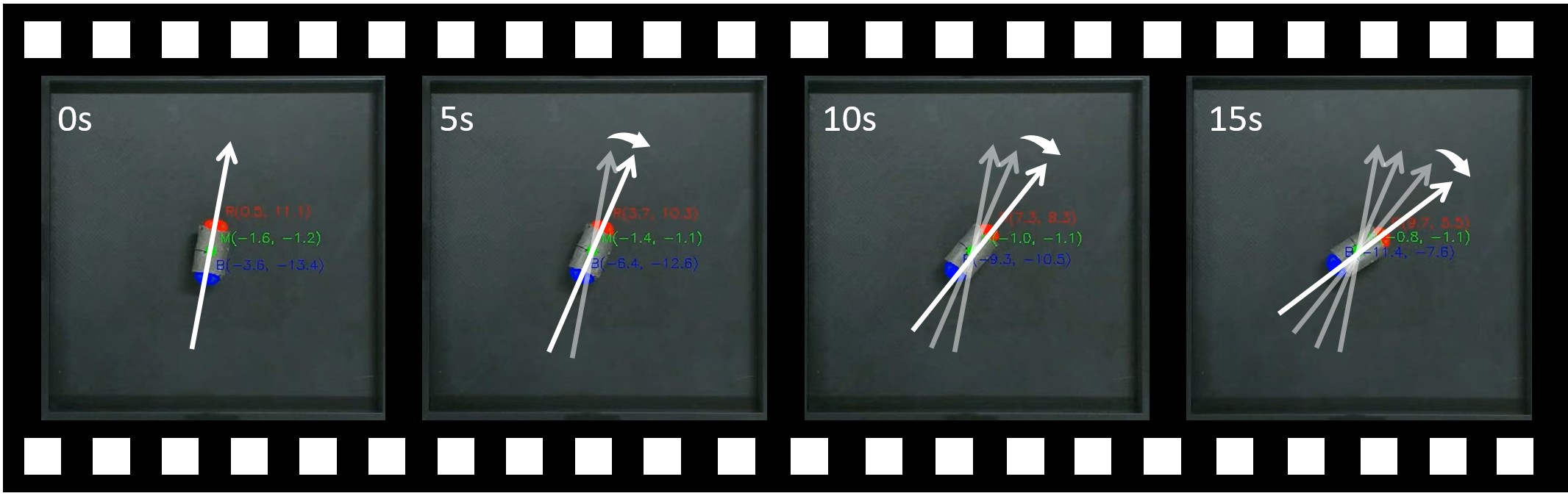}
    \caption{Capsule rotation sequence at 5-second intervals.}
    \label{fig:rotate_sequence}
\end{figure}

\subsection{Locomotion Under Challenging Surface Conditions}
The capsule robot's locomotion ability was further evaluated under three more surface conditions: a $7.5^{\circ}$ inclined silicone surface, a dry simulated stomach environment, and a wet simulated stomach environment.

\subsubsection{Inclined Silicone Surface}

Inclined Silicone Surface: On the $7.5^{\circ}$ inclined silicone surface, the capsule robot maintained a locomotion speed comparable to that on the flat PLA surface. However, to climb the slope, a greater effective magnetic torque was required to overcome gravitational resistance. This increased torque arises because the permanent magnet must rotate through a larger angle before the capsule realigns with the magnetic flux lines, introducing a delay that produces higher restoring torque. The robot successfully traversed the incline, nevertheless with a greater initial misalignment between the permanent magnet and the capsule was observed. The trajectory under this condition was shown in Fig.~14, with frame-by-frame positions extracted at 1-second intervals.

\begin{figure}[htbp]
    \centering
    \includegraphics[width=0.9\linewidth]{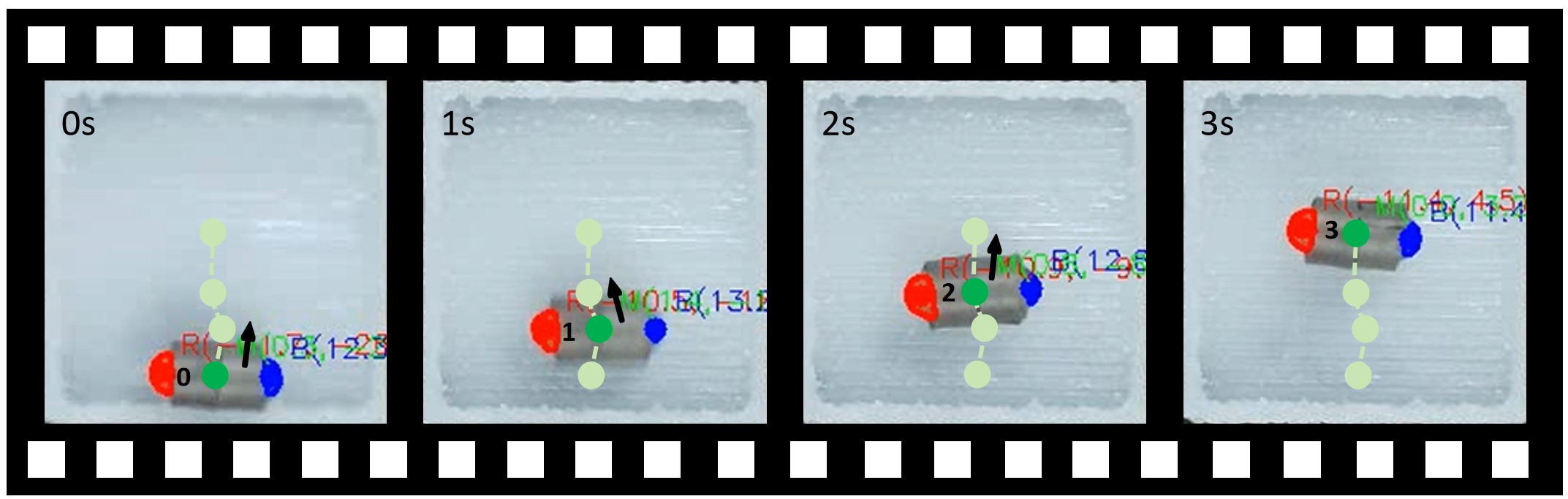}
    \caption{Capsule rolling sequence on $10^{\circ}$ inclined silicone surface at every 1-second intervals.}
    \label{fig:roll_incline}
\end{figure}
\subsubsection{Simulated Stomach (Dry Condition)}
\begin{figure}[htbp]
    \centering
    \includegraphics[width=0.95\linewidth]{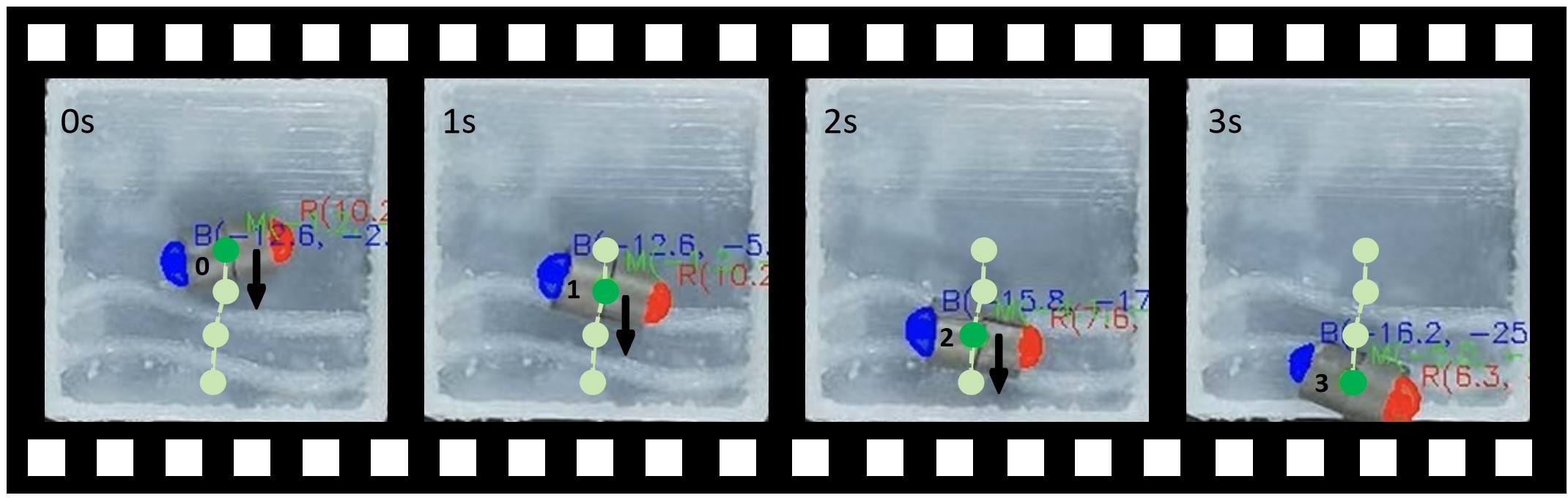}
    \caption{Capsule rolling sequence in dry simulated stomach condition at every 1-second intervals.}
    \label{fig:roll_stomach_dry}
\end{figure}

In the dry simulated stomach environment, the textured silicone introduced relatively high friction, which impeded the capsule’s ability to change direction. Although the robot maintained its speed, turning became less effective due to the adhesive interaction between the dry surfaces. Additionally, when traversing the 2~mm protrusions that simulate gastric folds, the capsule required a relatively large magnetic torque to overcome the resistance. The capsule was also discovered to traverse the 5~mm protrusion on the edge of the test base. While it was able to pass over the protrusions successfully, this process occasionally led to increased positional error due to sudden shifts in direction or momentum. A sequence of positions captured at 1-second intervals is shown in Fig.~\ref{fig:roll_stomach_dry}.

\subsubsection{Simulated Stomach (Wet Condition)}

\begin{figure}[htbp]
    \centering
    \includegraphics[width=0.95\linewidth]{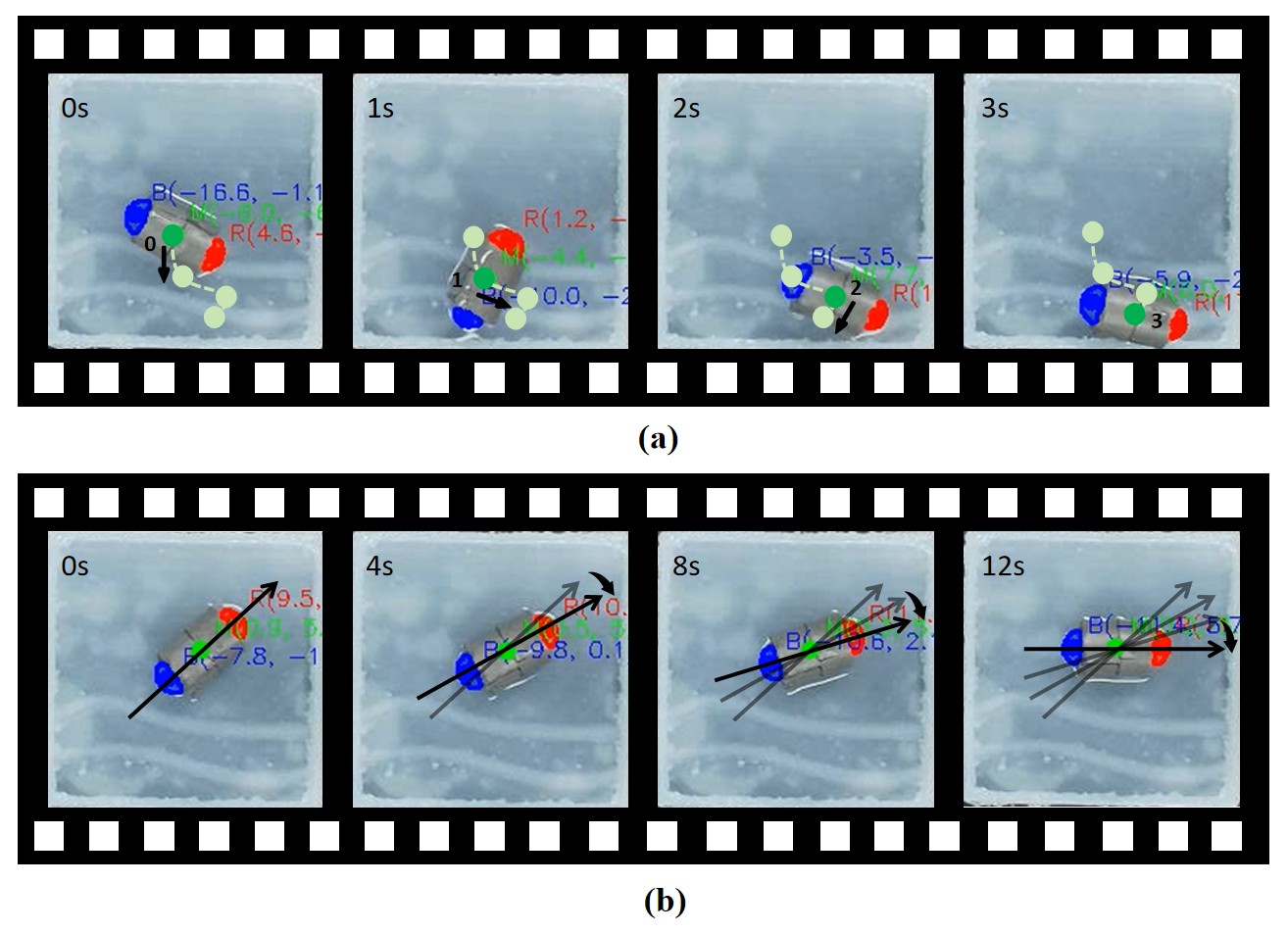}
    \caption{Movement of capsule under simulated stomach environment: (a) Capsule rolling sequence in wet simulated stomach condition at every 1-second intervals; (b) Capsule turning sequence in wet simulated stomach condition at every 4-second intervals.}
    \label{fig:roll_stomach_wet}
\end{figure}

When water was applied to the silicone surface to simulate a wet gastric environment, friction was significantly reduced. This allowed the capsule to change direction more smoothly compared to the dry condition. However, the lower friction also led to occasional slippage, increasing positioning error despite the improved maneuverability. In the rolling sequences (Fig. \ref{fig:roll_stomach_wet}(a)), we also observe noticeable yaw fluctuations; these are likely induced by surface texture that intermittently steer the capsule off its nominal path, compounding the slip effect. A practical consequence is that, after a slip, the internal magnetic dipole may need to rotate more than one nominal cycle before net forward translation resumes, creating a phase lag between commanded field rotation and displacement and thus violating our simplified rolling model. Future work will incorporate slip-aware, trial-and-error control, for example slip detection and phase/adaptive micro-corrections.

The capsule’s rolling behavior under wet conditions is shown in Fig.  \ref{fig:roll_stomach_wet}(a) with 1s frames; turning under the same condition is shown in Fig.  \ref{fig:roll_stomach_wet}(b) with 4s frames.


\vspace{-2mm}



\subsection{Error and Performance Summary}
The performance across the four test conditions was evaluated from overhead-camera trajectories using two error metrics. Because the raw trajectories exhibit heterogeneous sampling densities, we performed a density equalization step to harmonize the distributions across runs. We define
\begin{equation}
    RMS_{x} = \sqrt{\frac{1}{N}\sum_{i=1}^{N}(x_i-0)^2},
\end{equation}
which quantifies the deviation of the capsule along the $x$-axis when repeatedly traversing along $y$ direction, and
\begin{equation}
    RMS_{angle} = \sqrt{\frac{1}{N}\sum_{i=1}^{N}
    \left(\left|\tan^{-1}\!\left(\frac{\Delta y_i}{\Delta x_i}\right)-90^\circ\right|\right)^2 }.
\end{equation}
which measures the angular deviation from the ideal straight-line trajectory. Prior to computing this metric and plotting the angle traces, we applied a Hampel-style spike filter with a sliding window of 11 samples and a threshold of $n_\sigma = 1$, replacing flagged outliers with the local median value. This step attenuates isolated measurement spikes while preserving the underlying trajectory dynamics.

As summarized in Table~\ref{tab:error_summary}, the capsule yields the smallest normalized trajectory error on the smooth base. Notably, Fig.~17 shows that the motion range on the smooth PLA surface is substantially larger than on silicone; despite this conservative bias, the error on the smooth surface remains lower. We attribute this to the smoother, more uniform geometry and boundary conditions aligning better with our modeling assumptions (flat, rigid plane with approximately uniform friction), whereas the wet silicone with protrusions violates these assumptions and introduces unmodeled slip and compliance, increasing deviation.

\vspace{-2mm}
\begin{figure}[htbp]
    \centering
    \includegraphics[width=\linewidth]{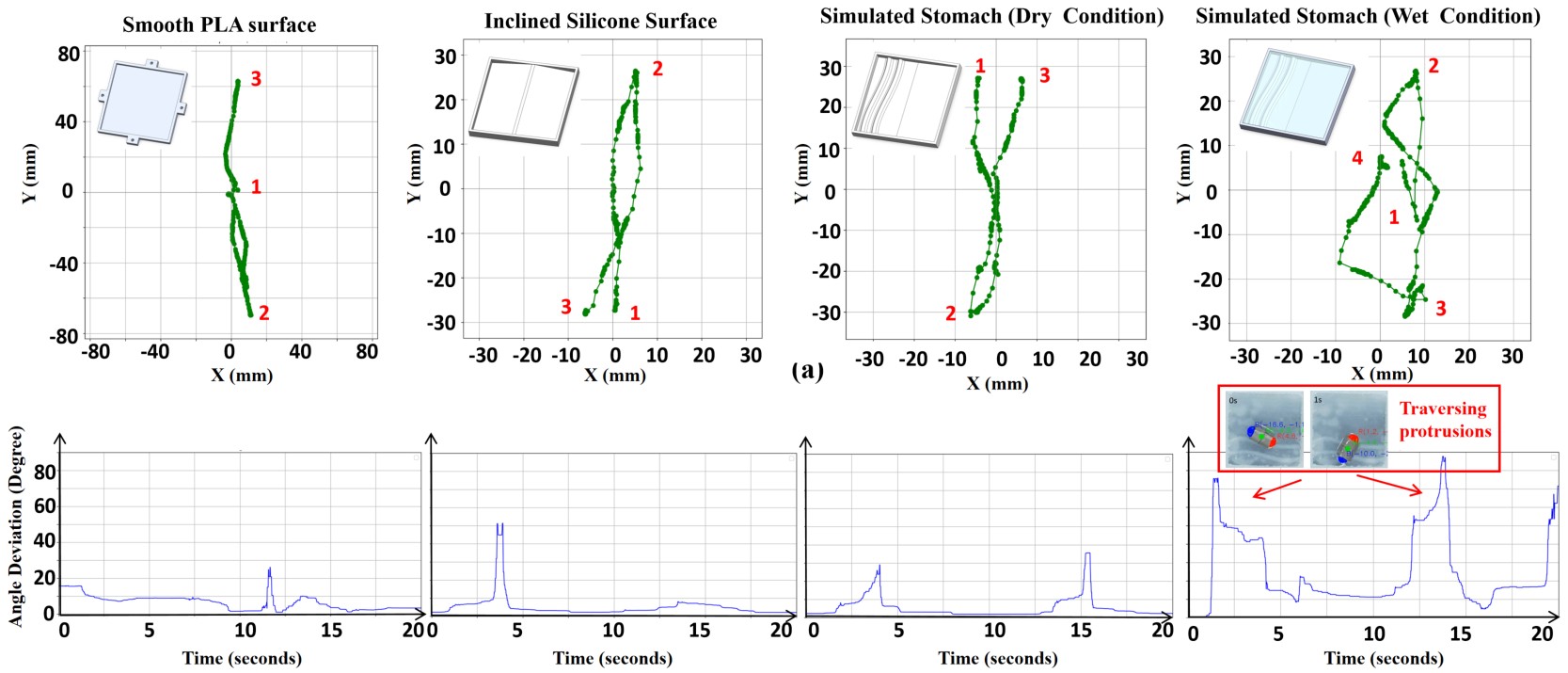}
    \caption{Result data analysis: (a) Capsule trajectories under four surface conditions, (b) Filtered Angle Deviation Over Time}
    \label{fig:trajectory}
\end{figure}

\begin{table}[htbp]
\centering
\begin{tabular}{|p{4.4cm}|p{1.2cm}|p{1.6cm}|}
\hline
\textbf{Base Type} & \textbf{$RMS_{x}$} & \textbf{$RMS_{angle}$} \\ \hline
\textbf{Smooth PLA Base} & 8.7095 & 3.5716 \\ \hline
\textbf{Silicone Base with Slope} & 14.8032 & 3.8039 \\ \hline
\textbf{Dry Silicone Base with Protrusions} & 15.5833 & 4.0662 \\ \hline
\textbf{Wet Silicone Base with Protrusions} & 32.0920 & 6.3050 \\ \hline
\end{tabular}
\caption{Quantitative summary of trajectory error for different surfaces.}
\label{tab:error_summary}
\end{table}

\section{Conclusion and Future Work}\label{sec_vi}
\vspace{-1mm}
This work introduced a compact capsule robot actuated by a 3D-printed anisotropic magnetic coating that replaces bulky internal permanent magnets. By programming magnetization during fabrication, the capsule achieves stable bi-directional rolling and smooth steering under rotating magnetic fields. Both simulation and experimental validation confirmed that the proposed NSSN/SNNS pole distribution provides reliable torque generation and strong anisotropy, enabling robust locomotion across smooth, inclined, and textured surfaces. In particular, the robot demonstrated stable rolling in simulated gastric environments under both dry and wet conditions, successfully overcoming frictional resistance and small protrusions. These results highlight the effectiveness of coating-based actuation in providing precise, continuous motion while preserving internal volume for future functional integration.

Future work can be broadly divided into two directions. \textbf{(i) Capsule optimization:} Beyond the current silicone–magnetic powder composite, further improvements can be achieved by adjusting the printing ratio of silicone and magnetic particles, exploring alternative materials, and optimizing the coating thickness and magnetic module layout. These refinements can balance objectives such as stronger magnetic force, lower thickness, suitable surface friction for different environments, and long-term robustness in gastric acid. In addition, embedding an inertial measurement unit (IMU) or other miniature sensors could provide onboard feedback to enhance control performance. \textbf{(ii) External control system optimization:} Building on the vision-assisted testbed, future efforts will focus on robotic-arm-controlled magnetic field generators with closed-loop feedback. While the current dynamic model assumes rolling on a continuous flat surface, real gastric walls present irregular terrain and disturbances that can reduce performance. To address this, learning-based control methods will be explored to adaptively manage complex contact conditions and ensure stable locomotion. Additional degrees of freedom, such as pitch control and stronger external fields, will further expand maneuverability and precision.

\section{Acknowledgement}
\label{sec: acknowledgement}
The authors would like to acknowledge AAA, for funding and support of this research.

\Urlmuskip=0mu plus 1mu\relax
\bibliographystyle{ieeetr}
\bibliography{ref}
\end{document}